\documentclass[sigconf]{acmart}

\usepackage{multirow}
\usepackage[normalem]{ulem}
\useunder{\uline}{\ul}{}

\AtBeginDocument{%
  }

\setcopyright{acmlicensed}
\copyrightyear{2025}
\acmYear{2025}
\acmDOI{XXXXXXX.XXXXXXX}
\acmConference[CIKM]{ACM International Conference on Information and Knowledge Management}{Nov 10--14, 2025}{COEX, SEOUL, KOREA}
\acmISBN{978-1-4503-XXXX-X/XXXX/XX}

\begin{document}

\title{TLCCSP: A Scalable Framework for Enhancing Time Series Forecasting with Time-Lagged Cross-Correlations}

\author{Jianfei Wu}  
\affiliation{%
  \institution{School of Artificial Intelligence \\ Beijing Normal University}  
  \city{Beijing}  
  \country{PR China}}  
\email{jianfeiwu@mail.bnu.edu.cn}  

\author{Wenmian Yang}  
\authornote{Corresponding authors.}
\affiliation{%
  \institution{Institute of Artificial Intelligence and Future Networks \\ Beijing Normal University}  
  \city{Zhuhai}  
  \country{PR China}}  
\email{wenmianyang@bnu.edu.cn}  

\author{Bingning Liu}  
\affiliation{%
  \institution{Department of Systems Science, Faculty of Arts and Sciences \\ Beijing Normal University}  
  \city{Zhuhai}  
  \country{PR China}}  
\email{liubingning@mail.bnu.edu.cn}

\author{Weijia Jia}  
\authornotemark[1]
\affiliation{%
  \institution{Institute of Artificial Intelligence and Future Networks, Beijing Normal University \\ Guangdong Key Lab of AI and Multi-Modal Data Processing, BNU-HKBU United International College}  
  \city{Zhuhai}  
  \country{PR China}}  
\email{jiawj@bnu.edu.cn} 

\renewcommand{\shortauthors}{Wu et al.}

\begin{abstract}
Time series forecasting is critical across various domains, such as weather, finance and real estate forecasting, as accurate forecasts support informed decision-making and risk mitigation. While recent deep learning models have improved predictive capabilities, they often overlook time-lagged cross-correlations between related sequences, which are crucial for capturing complex temporal relationships. To address this, we propose the Time-Lagged Cross-Correlations-based Sequence Prediction framework (TLCCSP), which enhances forecasting accuracy by effectively integrating time-lagged cross-correlated sequences. TLCCSP employs the Sequence Shifted Dynamic Time Warping (SSDTW) algorithm to capture lagged correlations and a contrastive learning-based encoder to efficiently approximate SSDTW distances. 

Experimental results on weather, finance and real estate time series datasets demonstrate the effectiveness of our framework. On the weather dataset, SSDTW reduces mean squared error (MSE) by 16.01\% compared with single-sequence methods, while the contrastive learning encoder (CLE) further decreases MSE by 17.88\%. On the stock dataset, SSDTW achieves a 9.95\% MSE reduction, and CLE reduces it by 6.13\%. For the real estate dataset, SSDTW and CLE reduce MSE by 21.29\% and 8.62\%, respectively. Additionally, the contrastive learning approach decreases SSDTW computational time by approximately 99\%, ensuring scalability and real-time applicability across multiple time series forecasting tasks.
\end{abstract}




\maketitle
\begin{figure*}[!ht]
  \centering
  {\includegraphics[width=0.95\textwidth]{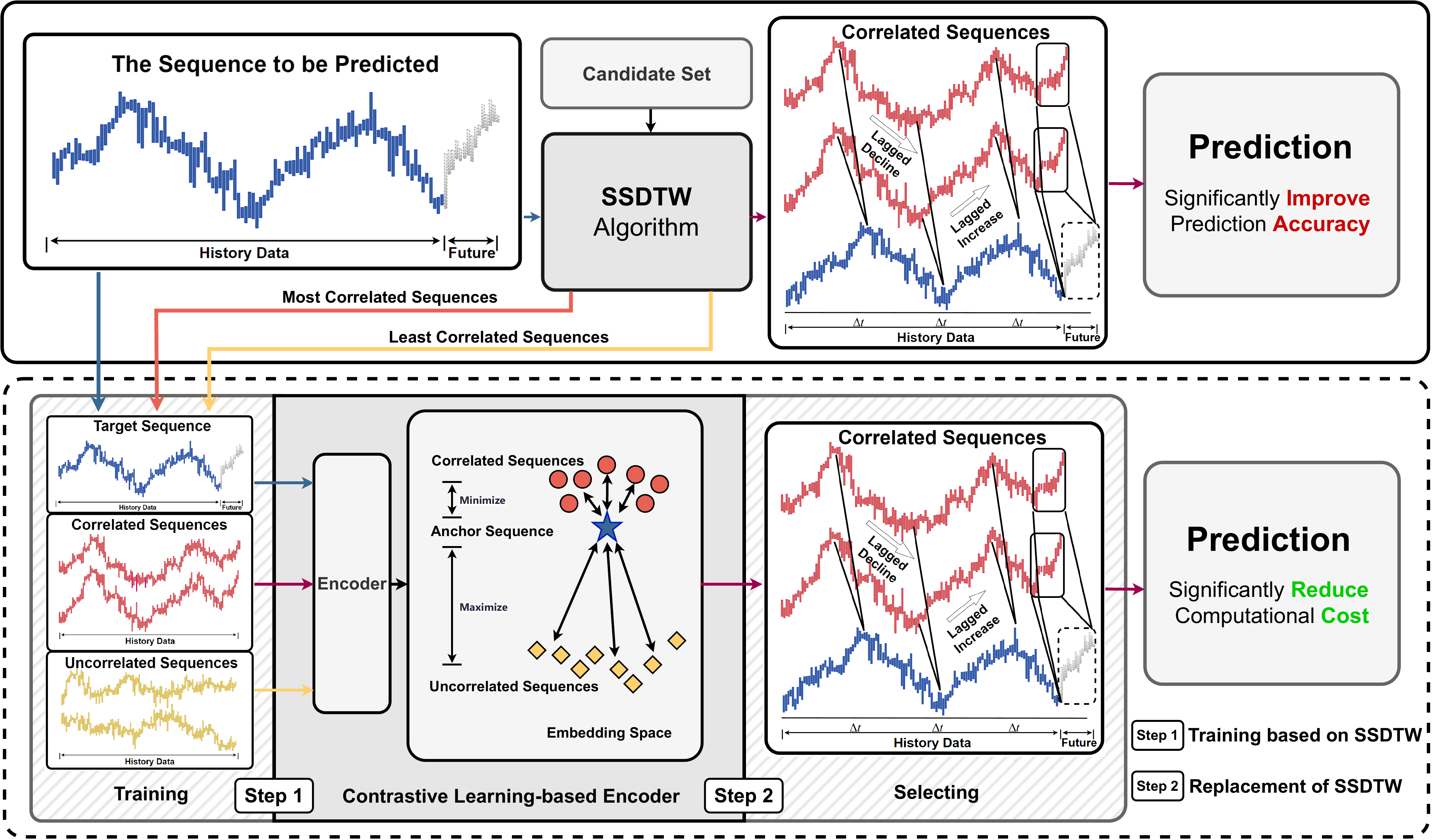}}
  \caption{Overall Framework of TLCCSP.}
  \label{fig:Overall}
\end{figure*}
\section{Introduction}
\label{sec:intro}
Time series forecasting, which predicts future values or trends based on historical data patterns, has been a central focus of research across various disciplines. Recent advancements in deep learning \cite{Transformer} have introduced powerful models \cite{Informer, TimesNet, iTransformer} capable of capturing intricate temporal dependencies and nonlinear relationships. These innovations have led to significant progress in applications such as meteorology, weather and finance. 

In complex forecasting scenarios such as temperature variation across different locations and stock price movements in financial markets, predictions are susceptible to a wide array of factors that extend beyond historical patterns alone. Recent research has demonstrated that incorporating auxiliary information can significantly enhance the accuracy of such complex time series forecasting tasks. For instance, Kevin et al. \cite{Weather_Aler-1} and Slater et al. \cite{Weather_Aler-2} discover that use auxiliary information like precipitation can improve the temperature forecasting accuracy. Billah et al. \cite{EmoFoca} show that integrating news sentiment data with deep learning models improves forecasting performance. Similarly, Tsai et al. \cite{CandleStick} report enhanced accuracy by combining the retrieval of analogous candlestick charts with predictive modeling. These findings highlight the crucial role of auxiliary information in developing robust and reliable prediction models, further emphasizing the importance of comprehensive data integration in time series forecasting.

In addition to the factors mentioned above, studies in weather forecasting and financial analysis \cite{ThunderTLCC, CrossCorr, TLCC} have identified time-lagged cross-correlations (TLCC) between individual sequences. For instance, temperature changes in cities along the same wind direction often exhibit sequential patterns, with upstream locations experiencing weather shifts that subsequently affect downstream areas. Similarly, trends in beverage company stocks frequently align with later movements in food company stocks. Despite the significance of this auxiliary information, it is often overlooked in deep learning-based prediction models.

Building on these observations, we propose the Time-Lagged Cross-Correlations based Sequence Prediction framework (TLCCSP) to improve the accuracy of existing time-series forecasting models. TLCCSP enhances predictions by automatically selecting time-lagged cross-correlated sequences as additional features. To identify these correlations, we introduce the Sequence Shifted Dynamic Time Warping (SSDTW) algorithm, an extension of the well-known Dynamic Time Warping (DTW) method \cite{DTWSpeech1, DTWSpeech2}. SSDTW adjusts sequences for time shifts before calculating distances, enabling the detection of lagged correlations in sequence movements. 

However, SSDTW incurs substantial computational overhead, making it impractical for applications requiring real-time analysis such as weather forecasting and financial market prediction. The computational complexity becomes particularly prohibitive when dealing with large-scale datasets featuring numerous time series, extensive historical data, and continuous updates. To address these challenges, we propose a contrastive learning-based encoder that maps time series sequences into a lower-dimensional embedding space. In this space, the distances between sequences closely approximate their SSDTW distances, thus dramatically reducing computational overhead. This approach enables real-time retrieval of time-lagged correlated sequences, making it feasible to perform real-time analysis. 

To validate the effectiveness of TLCCSP, we apply our proposed framework to multiple time-series forecasting models using three distinct datasets: a comprehensive weather dataset from the National Oceanic and Atmospheric Administration (NOAA) \cite{NOAA}, a large-scale stock market dataset \cite{Fnspid}, and a large-scale open-domain real estate price prediction dataset \cite{aaai2025retqa}. Experimental results demonstrate the effectiveness of our approach across all domains. 

The contributions of our study are summarized as follows:


\begin{itemize}
\item We propose TLCCSP, which enhances the accuracy of existing time-series forecasting models for prediction tasks with time-lagged dependencies by automatically identifying and leveraging cross-correlated sequences that exhibit temporal offsets as additional information.

\item We develop a contrastive learning-based encoder that significantly reduces the computational complexity of SSDTW while preserving its effectiveness. 

\item We apply our framework to multiple time-series forecasting models using three distinct datasets: a comprehensive weather dataset from NOAA\cite{NOAA}, a large-scale stock market dataset \cite{Fnspid}, and an open-domain real estate price prediction dataset \cite{aaai2025retqa}. Experimental results demonstrate that TLCC effectively enhances deep learning-based time series forecasting across domains with complex temporal dependencies.

\end{itemize} 

\section{Related Work}
In this section, we summarize general models for time series forecasting, then review advances in vertical domains such as weather, finance, and real estate.

\subsection{General Models for Time Series Forecasting}

Time series forecasting has been a fundamental research topic, with classical statistical approaches such as ARIMA \cite{AndersonKendall1976, PonceFloresEtAl2019} and Exponential Smoothing \cite{Gardner1985, CansuEtAl2024} providing well-established frameworks for capturing temporal dependencies. However, these traditional methods often face limitations when dealing with complex nonlinear dynamics commonly present in real-world data. In recent years, the emergence of deep learning methods has introduced new opportunities for addressing the complexities of time series data. Prominent examples include Convolutional Neural Networks (CNNs) \cite{WuEtAl2023}, Transformer-based models \cite{LimZohren2021, HaugsdalEtAl2023}, and TimesNet \cite{TimesNet}, which emphasizes feature extraction. Additionally, RNN-based architectures, such as Long Short-Term Memory (LSTM) networks \cite{Stefenon2024HypertunedWC, Fjellstrom2022Long} and the Long- and Short-term Time-series Network (LSTNet) \cite{WangEtAl2023}, have shown effectiveness in capturing both local and global patterns within time series data.

\subsection{Research on Time Series Forecasting in Vertical Domains}
In domain-specific applications such as weather forecasting, stock price prediction, and real estate price prediction, time series analysis faces additional challenges due to the complex nature of each field and the variety of factors affecting the data \cite{WeatherSurvey, Koval2023,realestate_survey}. 

In weather forecasting, studies such as Dong et al. \cite{DongXGBoost} and Zheng et al. \cite{ZhengShortRange} improved short-term precipitation forecasts by integrating machine learning with domain knowledge, while Bi et al. \cite{nature-3D} demonstrated that 3D deep networks can effectively balance accuracy and computational cost.
For stock price prediction, Hsu et al. \cite{Hsu2021} showed that incorporating industry data and semantic features from financial reports enhances predictive accuracy. More recent work has applied graph kernels \cite{GraphKernal} and graph learning \cite{GraphLearning} to capture complex relationships between price sequences.
In real estate, combining diverse data sources \cite{realestate_survey, realestate_fusion} and using spatiotemporal models such as CNN-LSTM \cite{TSestate} have led to notable improvements in price prediction.

Despite the advancements in these domains, existing deep learning models often fail to comprehensively address time-lagged cross-correlations (TLCC) across relevant sequences. Our proposed framework TLCCSP aims to explicitly model these complex interdependencies, enhancing the precision of time series forecasting in each domain.

\section{Methodology}


In this section, we introduce the TLCCSP framework in detail. Specifically, we provide the problem definition and overview in Section \ref{ov}. Then, we introduce the SSDTW algorithm in Section \ref{ssdtw}. Finally, we propose a contrastive learning-based encoder to reduce the computational complexity of SSDTW in Section \ref{cl}.
\subsection{Problem Definition and Framework Overview}
\label{ov}

As mentioned before, TLCCSP aims to enhance predictions by automatically selecting time-lagged cross-correlated time sequences as additional information.

We denote the historical data sequence of the target sequence $A = \{a_1, a_2, ..., a_T\}$  as input, where $T$ represents the number of time steps. Additionally, we further define the candidate sequence set $S = \{S_1, S_2, ..., S_N\}$, where $N$ is the total number of the candidate sequence, and each sequence $S_i \in S$ has the same data structure as $A$.

The overall framework is shown in Figure.~\ref{fig:Overall}, TLCCSP first calculates the TLCC between the target sequence $A$ and each candidate stock $S_i$ by our designed SSDTW (in Section \ref{ssdtw}):
\begin{equation}
\label{eq1}
d_{A,S_i} = SSDTW(A, S_i), \quad i \in \{1,2,...,N\},
\end{equation}

Then, a subset $S^* \subset S$ containing the top $K^s$ sequence with the highest TLCC values is selected as :
\begin{equation}
\label{eq2}
S^* = \{S_i | \text{rank}(d_{A,S_i}), i \in [1:K^s]\},
\end{equation}
where $\text{rank}(\cdot)$ denotes sort in in ascending order.

Finally, TLCCSP incorporates the sequence of target sequence $A$ and auxiliary sequence subset $S^*$ for future prediction as:
\begin{equation}
\label{eq3}
\hat{Y} = f(\{S_t\}_{t=1}^T, \{S_t^*\}_{t=1}^T),
\end{equation}
where $S_t^*$ represents the selected correlated sequences. 

The subsequent subsection details the computational procedure of the SSDTW algorithm and presents optimization strategies to address its high computational complexity.

\subsection{Sequence Shift Dynamic Time Warping algorithm (SSDTW)}
\label{ssdtw}

In real-world scenarios, the lag time between different sequence can vary significantly, ranging from a few days to several weeks \cite{TLCC}. To capture these varying time lags, SSDTW is specifically designed to calculate sequence similarity across different time shifts, effectively modeling the inherent heterogeneity present in real-world environments.

Formally, given a target sequence $A$ and a candidate sequence  $S$ over time horizon $T$, we first define a set of time shift parameters $\mathcal{T} = \{\tau_1, \tau_2, \dots, \tau_c\}$, where $\tau_i$ represents the $i$-th shift window in days with $\tau_1 < \tau_2 < \dots < \tau_c$ to capture different temporal scales, and $c$ denotes the total number of shift windows. For each $\tau \in \mathcal{T}$, we apply a temporal shift operation to the candidate sequence $S$: 

\begin{equation}  
\label{eq4}
\begin{aligned}  
S_{\tau} &= \text{TimeShift}(S, \tau) \\
&= \{s_{1}, s_{2}, ..., s_{T-\tau}\}, \quad \tau \in \mathcal{T}  
\end{aligned}  
\end{equation}  

The SSDTW distance between sequences $A$ and $S$ is then computed as the minimum DTW distance across all possible time shifts:          
\begin{equation}  
\label{eq5}
\text{SSDTW}(A, S) = \min_{\tau \in \mathcal{T}} \text{DTW}(A, S_{\tau}),
\end{equation}
where $DTW(\cdot)$ is calculated by the Dynamic Time Warping \cite{DTW2020,DTWSurvey} algorithm that widely used for measuring time series similarity.

\subsection{Contrastive Learning-based Encoder}
\label{cl}
Although TLCCSP could improve prediction accuracy, the computational complexity of SSDTW algorithm is too high for the real-world appliations. For example, in a market with $N$ stocks and $T$ time steps, calculating the TLCC for all possible stock pairs results in a total time complexity of $O(N^2 \cdot T^2)$,  which presents significant computational challenges, particularly when handling large historical price datasets spanning thousands of trading days. Furthermore, since the algorithm is based on dynamic programming, it is not well-suited for parallel computation. 

To address these limitations, we propose an ingenious contrastive learning-based encoder that maps time sequence data into an embedding space. In this space, the distances between sequences closely approximate their SSDTW distances. 


Specifically, given a target sequence $A$, we aimes to minimize distance between the target sequence and its correlated sequences while simultaneously maximizing the distance from uncorrelated sequences in the embedding space. Therefore, we define the positive sample set $\mathcal{P}(A)$ and negative sample set $\mathcal{N}(A)$ as: 
\begin{equation}  
\begin{aligned}  
\label{eq6}
\mathcal{P}(A) &= \{S_i \in \mathcal{S} \mid \text{rank}(\text{SSDTW}(A, S_i)), i \leq K^e\}, \\
\mathcal{N}(A) &= \{S_j \in \mathcal{S} \mid \text{rank}(\text{SSDTW}(A, S_j)), i >N-K^e\}, 
\end{aligned}  
\end{equation}
where the most and least correlated $K^e$ sequences are selected as positive samples and negative samples, respectively.

\begin{table*}[!ht]
\centering
\caption{The Performance comparison of different methods on various models, with the best results in \textbf{bold} and the second-best results \underline{underlined}}.
\fontsize{9}{10}\selectfont 
\renewcommand{\arraystretch}{1.5} 
\resizebox{\textwidth}{!}{
\begin{tabular}{c|l|ccccc|ccccc}
\hline
\textbf{Dataset}                                                                & \textbf{Model}        & \multicolumn{5}{c|}{\textbf{MSE}}                                                                                                               & \multicolumn{5}{c}{\textbf{MAE}}                                                                                                               \\ \cline{3-12} 
\multicolumn{1}{l|}{}                                                           &                       & \multicolumn{1}{l}{Single} & \multicolumn{1}{l}{SSDTW} & \multicolumn{1}{l}{$\Delta$} & \multicolumn{1}{l}{CLE} & \multicolumn{1}{l|}{$\Delta$} & \multicolumn{1}{l}{Single} & \multicolumn{1}{l}{SSDTW} & \multicolumn{1}{l}{$\Delta$} & \multicolumn{1}{l}{CLE} & \multicolumn{1}{l}{$\Delta$} \\ \hline
\multirow{8}{*}{\textbf{\begin{tabular}[c]{@{}c@{}}Weather\\ $(\times 10^{-2})$\end{tabular}}} & \textbf{CNN}          & 16.52                      & \textbf{14.62}            & -11.48\%                     & {\ul 15.55}             & -5.90\%                       & 29.94                      & {\ul 27.79}               & -7.18\%                      & \textbf{27.75}          & -7.32\%                      \\
                                                                                & \textbf{RNN}          & 16.17                      & {\ul 15.80}               & -2.29\%                      & \textbf{14.85}          & -8.20\%                       & 29.13                      & {\ul 28.52}               & -2.11\%                      & \textbf{27.78}          & -4.66\%                      \\
                                                                                & \textbf{LSTM}         & 14.61                      & {\ul 11.94}               & -18.28\%                     & \textbf{11.73}          & -19.67\%                      & 27.63                      & \textbf{24.16}            & -12.56\%                     & {\ul 24.64}             & -10.80\%                     \\
                                                                                & \textbf{Transformer}  & 13.11                      & {\ul 11.44}               & -12.73\%                     & \textbf{9.98}           & -23.88\%                      & 25.95                      & {\ul 23.57}               & -9.19\%                      & \textbf{22.63}          & -12.80\%                     \\
                                                                                & \textbf{Informer}     & 13.38                      & \textbf{11.42}            & -14.61\%                     & {\ul 12.66}             & -5.39\%                       & 26.58                      & \textbf{23.56}            & -11.35\%                     & {\ul 25.78}             & -2.99\%                      \\
                                                                                & \textbf{iTransformer} & 15.04                      & {\ul 10.92}               & -27.42\%                     & \textbf{10.37}          & -31.05\%                      & 27.70                      & \textbf{22.97}            & -17.07\%                     & {\ul 23.12}             & -16.53\%                     \\
                                                                                & \textbf{TimesNet}     & 14.31                      & {\ul 10.48}               & -26.75\%                     & \textbf{9.57}           & -33.15\%                      & 27.06                      & {\ul 23.11}               & -14.60\%                     & \textbf{20.33}          & -24.88\%                     \\ \cline{2-12} 
                                                                                & \textbf{Average}      & 14.73                      & {\ul 12.38}               & -16.01\%                     & \textbf{12.10}          & -17.88\%                      & 27.71                      & {\ul 24.81}               & -10.47\%                     & \textbf{24.58}          & -11.32\%                     \\ \hline
\multirow{8}{*}{\textbf{\begin{tabular}[c]{@{}c@{}}Stock\\ $(\times 10^{-3})$\end{tabular}}}   & \textbf{CNN}          & 3.19                       & {\ul 2.99}                & -6.27\%                      & \textbf{2.81}           & -11.91\%                      & 40.59                      & \textbf{37.05}            & -8.72\%                      & {\ul 39.70}             & -2.19\%                      \\
                                                                                & \textbf{RNN}          & 2.43                       & \textbf{2.17}             & -10.70\%                     & {\ul 2.36}              & -2.88\%                       & 31.95                      & {\ul 26.19}               & -18.03\%                     & \textbf{25.18}          & -21.19\%                     \\
                                                                                & \textbf{LSTM}         & {\ul 1.65}                 & 1.68                      & 1.82\%                       & \textbf{1.60}           & -3.03\%                       & 19.67                      & {\ul 15.16}               & -22.93\%                     & \textbf{14.56}          & -25.98\%                     \\
                                                                                & \textbf{Transformer}  & 0.75                       & \textbf{0.67}             & -10.67\%                     & {\ul 0.69}              & -8.00\%                       & 14.99                      & \textbf{11.83}            & -21.08\%                     & {\ul 13.04}             & -13.01\%                     \\
                                                                                & \textbf{Informer}     & {\ul 2.81}                 & \textbf{2.48}             & -11.74\%                     & 2.97                    & 5.69\%                        & 36.22                      & \textbf{31.20}            & -13.86\%                     & {\ul 35.00}             & -3.37\%                      \\
                                                                                & \textbf{iTransformer} & 0.84                       & \textbf{0.62}             & -26.19\%                     & {\ul 0.63}              & -25.00\%                      & 15.22                      & {\ul 14.70}               & -3.42\%                      & \textbf{14.68}          & -3.55\%                      \\
                                                                                & \textbf{TimesNet}     & 0.89                       & \textbf{0.70}             & -21.35\%                     & {\ul 0.73}              & -17.98\%                      & 21.82                      & \textbf{15.62}            & -28.41\%                     & {\ul 15.62}             & -28.41\%                     \\ \cline{2-12} 
                                                                                & \textbf{Average}      & 1.79                       & \textbf{1.62}             & -9.95\%                      & {\ul 1.68}              & -6.13\%                       & 25.78                      & \textbf{21.68}            & -15.91\%                     & {\ul 22.54}             & -12.57\%                     \\ \hline
\multirow{8}{*}{\textbf{Real Estate}}                                           & \textbf{CNN}          & 1.37                       & \textbf{0.93}             & -32.22\%                     & {\ul 1.24}              & -9.49\%                       & 0.92                       & \textbf{0.79}             & -14.01\%                     & {\ul 0.88}              & -4.35\%                      \\
                                                                                & \textbf{RNN}          & 1.43                       & \textbf{1.28}             & -10.49\%                     & {\ul 1.36}              & -4.90\%                       & 0.93                       & \textbf{0.91}             & -2.15\%                      & {\ul 0.92}              & -1.08\%                      \\
                                                                                & \textbf{LSTM}         & 1.54                       & \textbf{0.97}             & -36.93\%                     & {\ul 1.43}              & -7.14\%                       & 0.95                       & \textbf{0.81}             & -14.28\%                     & {\ul 0.96}              & 1.05\%                       \\
                                                                                & \textbf{Transformer}  & {\ul 1.55}                 & \textbf{1.15}             & -25.81\%                     & 1.57                    & 1.29\%                        & 1.11                       & \textbf{0.85}             & -23.29\%                     & {\ul 0.99}              & -10.65\%                     \\
                                                                                & \textbf{Informer}     & 1.42                       & \textbf{1.08}             & -23.94\%                     & \textbf{1.08}           & -23.94\%                      & 1.00                       & {\ul 0.84}                & -16.33\%                     & \textbf{0.83}           & -16.83\%                     \\
                                                                                & \textbf{iTransformer} & 0.87                       & {\ul 0.83}                & -4.30\%                      & \textbf{0.81}           & -6.90\%                       & 0.76                       & {\ul 0.74}                & -2.46\%                      & \textbf{0.71}           & -6.58\%                      \\
                                                                                & \textbf{TimesNet}     & {\ul 0.87}                 & 0.88                      & 1.28\%                       & \textbf{0.78}           & -10.34\%                      & 0.76                       & {\ul 0.77}                & 0.99\%                       & \textbf{0.68}           & -10.53\%                     \\ \cline{2-12} 
                                                                                & \textbf{Average}      & 1.29                       & \textbf{1.02}             & -21.29\%                     & {\ul 1.18}              & -8.62\%                       & 0.92                       & {\ul 0.82}                & -11.15\%                     & \textbf{0.85}           & -7.10\%                      \\ \hline
\end{tabular}
}
\label{tab:performance}
\end{table*}

With the positive and negative samples, the contrastive learning objective can then be formulated as:  
\begin{equation}  
\label{eq7}
    Z_{pos} = \sum_{S^+ \in \mathcal{P}(A)} \exp\left(\frac{\text{sim}(\mathcal{E}(A), \mathcal{E}(S^+))}{\lambda}\right),
\end{equation}  
\begin{equation}  
\label{eq8}
    Z_{neg} = \sum_{S^- \in \mathcal{N}(A)} \exp\left(\frac{\text{sim}(\mathcal{E}(A), \mathcal{E}(S^-))}{\lambda}\right), 
\end{equation}  
\begin{equation}  
    \mathcal{L} = -\log\left(\frac{Z_{pos}}{Z_{pos} + Z_{neg}}\right),
\end{equation}  
where $\mathcal{E}(\cdot)$ denotes the encoder that maps sequences to the embedding space, $\text{sim}(\cdot,\cdot)$ represents the cosine similarity between embedded representations, and $\lambda$ is a temperature parameter controlling the concentration level of the distribution. In particular, the encoder is a 3-block convolutional neural network as proposed in \cite{adatime}.

With the aid of the well-trained encoder, correlations among all sequences can be efficiently determined by projecting them into an embedding space and measuring the distances between their embeddings, thereby eliminating the need for the SSDTW algorithm.

Compared with the SSDTW algorithm, the contrastive learning-based encoder substantially reduces the computational cost of calculating correlations.

\section{Experiments}

In this section, we begin by introducing the datasets, settings, and baselines employed in our experiments. Subsequently, we conduct an analysis of the experimental results of the baseline methods in conjunction with the TLCCSPP framework. Finally, we perform an ablation study to validate the effectiveness of our designed framework.

\subsection{Dataset and Experimental Setup}  


To evaluate the effectiveness of our proposed framework, we conducted comprehensive experiments using three different datasets. Specifically, we performed daily temperature predictions using the weather dataset published by NOAA\cite{NOAA}, daily stock price predictions on the FNSPID dataset \cite{Fnspid}, and monthly real estate price predictions using the RETQA dataset \cite{aaai2025retqa}.

\subsubsection{Weather Dataset}
The NOAA weather dataset is released by the National Oceanic and Atmospheric Administration (NOAA), provides comprehensive and high-quality meteorological observations collected from thousands of stations worldwide. Although the dataset contains various meteorological features, we use only the daily average temperature in our experiments. Following the settings in FNSPID \cite{Fnspid}, we utilize 49 days of historical data to forecast the target values for the subsequent 3 days. we employ daily average temperature data from 1,882 cities without missing values, covering 364 days in 2019, to calculate SSDTW distances and construct candidate correlated stock sets. These distances guide the training of the contrastive learning-based encoder.

For the time series forecasting task, we use daily average temperature data from 2020 and follow the sliding window approach from FNSPID \cite{Fnspid}(e.g., [1,50], [2,51], ..., [T-49,T]). The first 49 days serve as inputs to predict the target at the 50th day. All temperature data are standardized using Z-score normalization. The 2020 dataset is randomly divided into training, validation, and test sets in a 6:2:2 ratio.

\subsubsection{Stock Dataset}
The FNSPID dataset \cite{Fnspid} is a publicly available large-scale financial time series benchmark that provides comprehensive stock trading records across multiple features. For this dataset, we use daily average prices of 4,426 stocks with continuous trading records spanning over 700 days before December 31, 2018, to compute SSDTW distances and train the contrastive learning-based encoder. Importantly, stocks included in the candidate pool exclude any stocks in the test set to maintain evaluation integrity and the selection of the test set follows the same in FNSPID. Unlike the weather dataset, we adopt the same settings as FNSPID by using three input features—daily trading volume, daily return rate, and daily average price—to predict the future daily return rate of stocks.

During the training phase, consistent with FNSPID settings \cite{Fnspid}, a sliding window of 50 days segments each historical sequence into overlapping windows. The first 49 days serve as inputs to predict the target at the 50th day. 

\subsubsection{Real Estate Dataset}
The RETQA dataset \cite{aaai2025retqa} is the first large-scale, open-domain Chinese tabular question answering dataset focusing on the real estate sector. Due to excessive missing values, we first apply TimesNet \cite{TimesNet}, a state-of-the-art time series imputation model with default hyperparameters, to fill all missing numerical entries. We then use monthly average sales price data from 14,086 plots in 2022 to calculate SSDTW correlations and train CLE models, followed by prediction on the 2023 monthly sales prices. Concretely, monthly average prices for the first nine months of 2023 serve as inputs for autoregressive forecasting of the last three months’ real estate prices in 2023.

All prediction-related features are standardized via Z-score normalization before training, and the dataset is randomly partitioned into training, validation, and test sets in a 6:2:2 ratio.


\begin{figure*}[!ht]
  \centering
  {\includegraphics[width=1\textwidth]{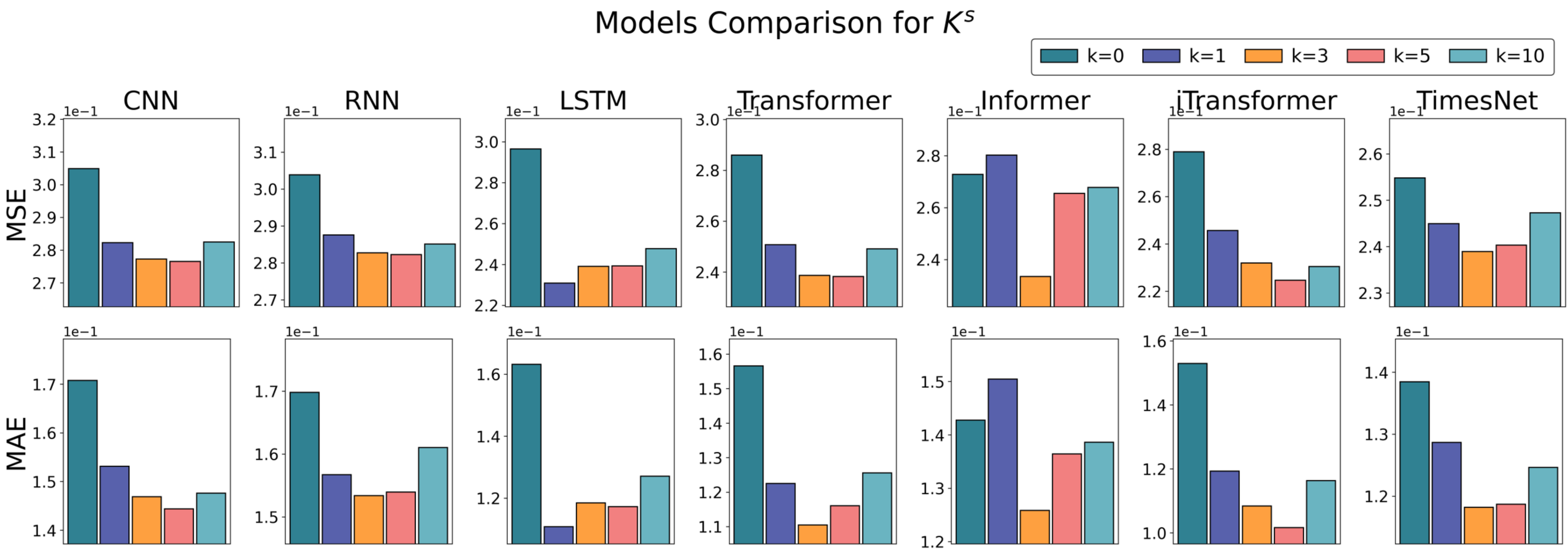}}
  \caption{MSE and MAE of Different models on weather dataset with different settings of $K^s$}
  \label{fig:Sen_CNT}
\end{figure*}
\begin{figure*}[!ht]
  \centering
  {\includegraphics[width=1\textwidth]{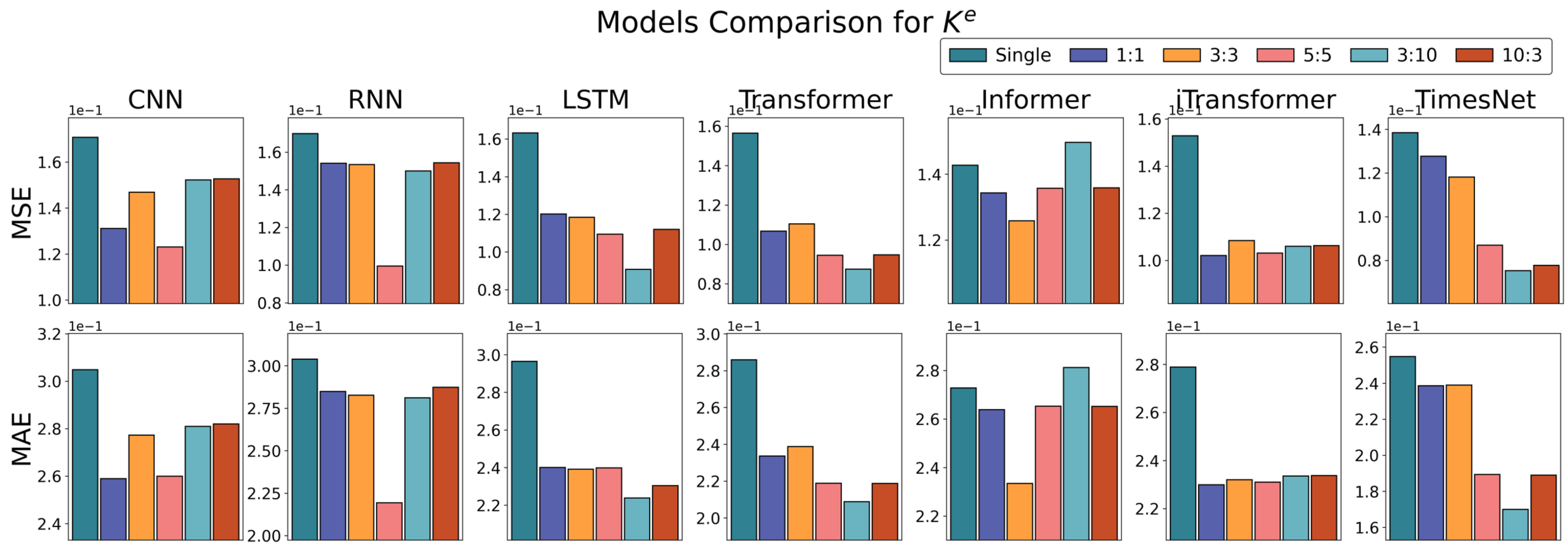}}
  \caption{MSE and MAE of Different models on weather dataset with different settings of $K^e$}
  \label{fig:Sen_posneg}
\end{figure*}

\subsection{Baseline}

To comprehensively evaluate the performance of our proposed framework, we design three methodological approaches for prediction:
\begin{itemize}  
    \item \textbf{Single}-series baseline: A conventional approach that relies solely on individual sequence's historical data, serving as a fundamental benchmark for performance comparison.
    
    \item \textbf{SSDTW}-based selection: An enhanced method that leverages the SSDTW algorithm to identify and select correlated reference sequences.  
    
    \item \textbf{C}ontrastive \textbf{L}earning-based \textbf{E}ncoder selection(\textbf{CLE}): An alternative and computationally efficient method for reference sequences selection.
\end{itemize}

To validate the robustness and versatility of our proposed framework, we conduct extensive experiments across multiple models, i.e., four basic models CNN \cite{CNN}, RNN \cite{RNN}, LSTM \cite{LSTM}, and Transformer \cite{Transformer}, and three state-of-the-art models TimesNet \cite{TimesNet}, Informer \cite{Informer}, and iTransformer \cite{iTransformer}. 
 
\subsection{Settings}
In real-world scenarios, the lag time between different sequences can vary significantly, ranging from a few days to several weeks \cite{TLCC}. To comprehensively capture these diverse temporal variations, we select different sets of shift windows \(\tau\) tailored to each dataset: \(\{1, 3, 5, 10\}\) days for the weather dataset, \(\{5, 10, 20, 30\}\) days for the stock price dataset, and \(\{1, 2, 3\}\) months for the real estate price dataset in equations Eq.\eqref{eq4} and Eq.\eqref{eq5}.

Following previous work, we maintain consistency in the network architecture settings for CNN, RNN, LSTM, Transformer, and TimesNet models in alignment with \cite{Fnspid}. The parameters for Informer \cite{Informer} and iTransformer \cite{iTransformer} are also adopted from their respective original papers. However, the training hyperparameters are set differently from those in the referenced work. 

For the weather and stock datasets, traditional models including CNN, RNN, and LSTM are trained with a learning rate of $1.0 \times 10^{-3}$, dropout rate of 0.2, batch size of 32, a maximum of 50 epochs, and early stopping with a patience of 5. Transformer-based models (Transformer, Informer, iTransformer, and TimesNet) used a lower learning rate of $1.0 \times 10^{-5}$, the same dropout rate of 0.2, batch size of 64, 50 epochs, and early stopping patience of 3. 
For the real estate dataset, the dropout rate is increased to 0.3; traditional models use a batch size of 16, while Transformer-based models use a batch size of 32. The maximum epochs and early stopping criteria remain consistent across all datasets and models. This hyperparameter setup ensure stable and effective training across different model architectures and datasets.

For other hyperparameters, $K^s$ in Eq. \eqref{eq2} is set as 3 ,  $K^e$ in Eq.\eqref{eq6} is set as 5, and $\lambda$ in Eq.\eqref{eq7} and Eq.\eqref{eq8} is 0.2, according to the optimal performance in the validation set. A more detailed sensitivity analysis is shown in Section \ref{sa}.

\begin{figure*}[!ht]
  \centering
  {\includegraphics[width=1\textwidth]{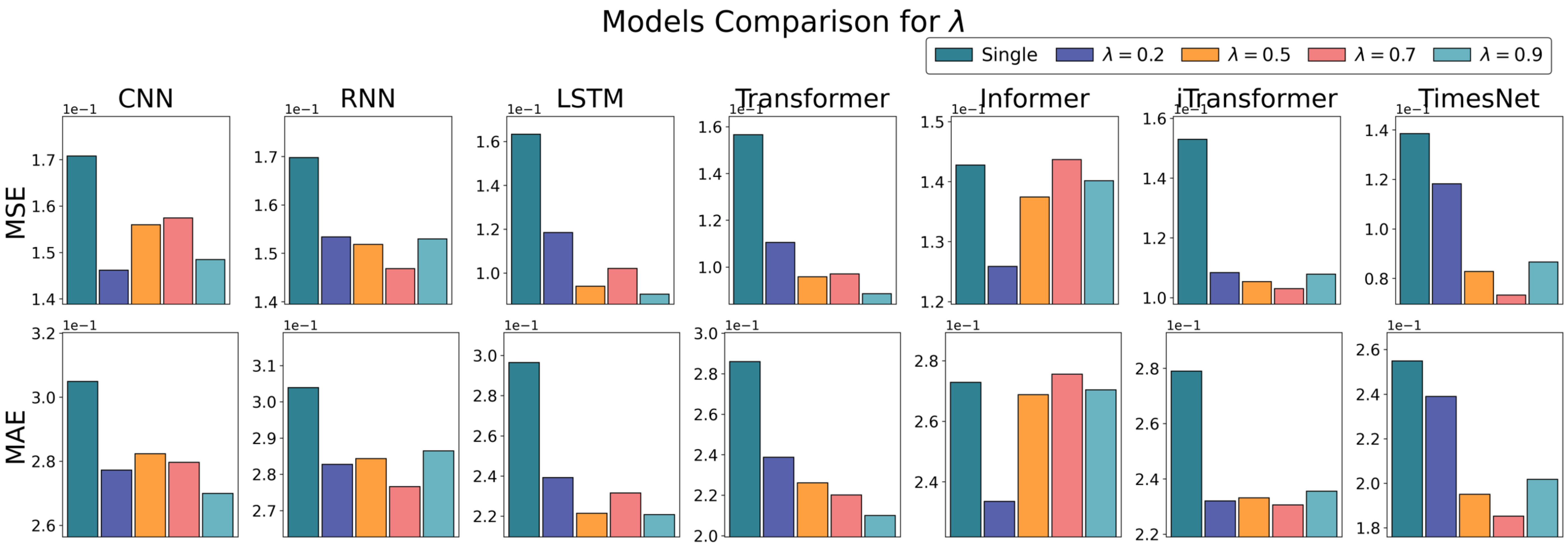}}
  \caption{MSE and MAE of Different models on weather dataset with different settings of $\lambda$}
  \label{fig:Sen_lambda}
\end{figure*}
\subsection{Result Analysis}


Based on the above experimental settings, we evaluate the performance of the TLCCSP framework across several baselines using Mean Squared Error (MSE) and Mean Absolute Error (MAE) as the assessment metrics. The comprehensive results across three diverse datasets, i.e., weather, stock price, and real estate, are summarized in Table.~\ref{tab:performance}.


The results show that both SSDTW and the Contrastive Learning Encoder (CLE) outperform the Single-Series Baseline models across all datasets.

Specifically, SSDTW achieves substantial improvements in prediction accuracy. For the weather dataset, SSDTW reduces MSE by up to 27.42\% and MAE by up to 17.07\%. Similarly, in the stock price dataset, SSDTW yields MSE reductions reaching 26.19\% and MAE reductions up to 28.41\%, reflecting its ability to capture meaningful temporal correlations. For the real estate dataset, SSDTW attains notable MSE reductions as high as 36.93\% and MAE improvements near 23.29\%, despite the dataset’s inherent challenges such as missing values.

Similarly, the CLE approach yields substantial gains by learning discriminative representations from time-lagged cross-correlated sequences. For example, CLE reduces MSE by up to 33.15\% and MAE by nearly 24.88\% on the weather dataset, and achieves MAE reductions of up to 25.98\% on the stock dataset. 

Across different backbone models (CNN, RNN, LSTM, Transformer, Informer, iTransformer, and TimesNet), both SSDTW and CLE consistently rank first or second. These findings underscore the effectiveness of incorporating time-lagged cross-correlated stock sequences to enhance stock price prediction performance.

We also observe that, although CLE serves as an approximation of the SSDTW correlation calculation, the predictive accuracy achieved by leveraging correlated sequences obtained via CLE is not necessarily inferior to those obtained through the SSDTW algorithm. This phenomenon may be attributed to the encoder in CLE, which is capable of capturing additional information beyond what is provided strictly by SSDTW correlations. Such supplementary representations likely contribute effectively to forecasting future trends in time series data.

In terms of computational efficiency, experiments are conducted in an Intel i9-14900K CPU and Nvidia RTX 4090 GPU environment. On the stock dataset, using the SSDTW algorithm to calculate TLCCs between all pairs of 4426 stocks requires 980 hours. In contrast, the well-trained encoder can compute pairwise stock similarities in just 3 hours, dramatically reducing computational overhead and enabling real-time financial analysis.

\subsection{Ablation Study}
\label{sa}

\begin{table}[h]
\centering
\caption{Ablation Study: MSE of Different Approaches on Stock Dataset}
\fontsize{9}{11}\selectfont
\begin{tabular}{l|cccc}
\hline
Model        & \multicolumn{1}{l}{Single} & \multicolumn{1}{l}{RS} & \multicolumn{1}{l}{DTW} & \multicolumn{1}{l}{SSDTW} \\ \hline
CNN          & 3.19                       & 3.57                   & 3.11                    & \textbf{2.99}             \\
RNN          & 2.43                       & 2.51                   & 2.39                    & \textbf{2.17}             \\
LSTM         & 1.65                       & 1.72                   & 1.68                    & \textbf{1.68}             \\
Transformer  & 0.75                       & 0.87                   & 0.69                    & \textbf{0.67}             \\
Informer     & 2.81                       & 2.73                   & \textbf{2.36}                    & 2.48             \\
iTransformer & 0.84                       & 1.12                   & 0.91                    & \textbf{0.62}             \\
TimesNet     & 0.89                       & 0.85                   & \textbf{0.69}           & 0.70                      \\ \hline
\end{tabular}
\label{AblationStudy}
\end{table}
\begin{figure*}[!ht]
  \centering
  {\includegraphics[width=1\textwidth]{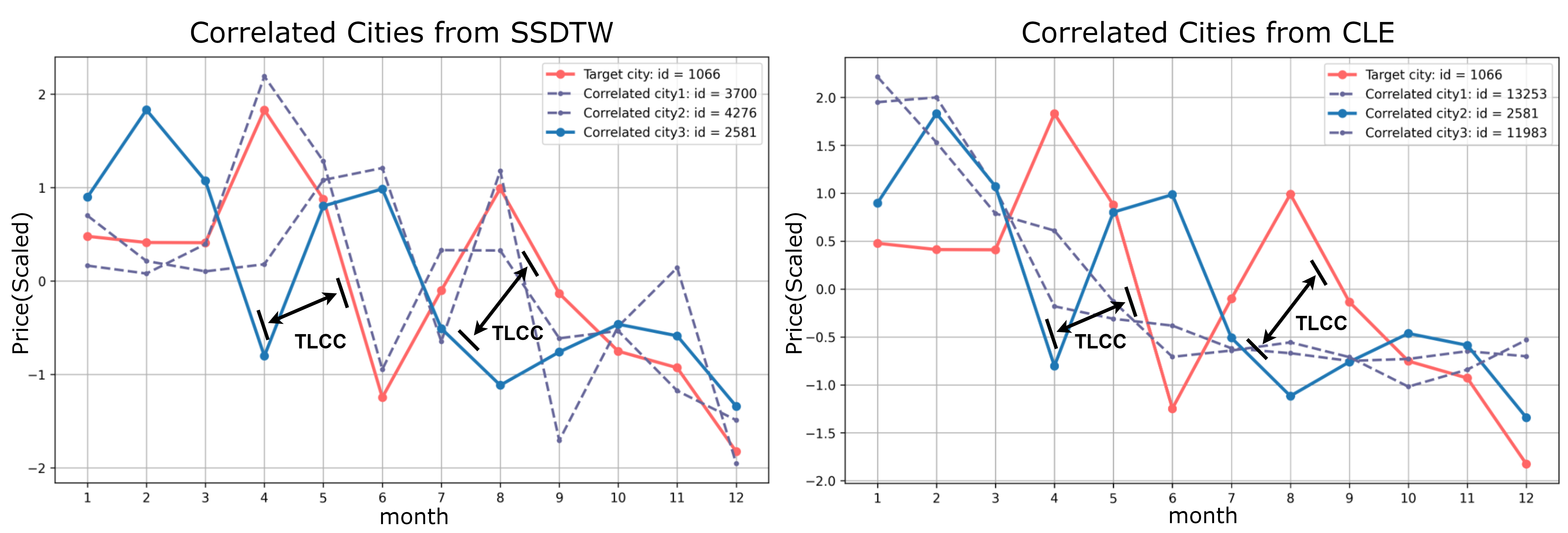}}
  \caption{Case Study: Correlated Sequences Selected by SSDTW and CLE}
  \label{fig:case-study}
\end{figure*}
To validate the effectiveness of using TLCC sequences as auxiliary information, we conduct a comprehensive comparative analysis on the dataset with the longest time series data, i.e., the stock price dataset. Specifically, we evaluate the following approaches for selecting correlated stocks:  
\begin{itemize}  
    \item Random Selection (RS): Stocks are selected randomly as auxiliary series.  
    \item DTW Algorithm: Correlated stocks are selected using the traditional dynamic time warping (DTW) algorithm.  
    \item SSDTW Algorithm: Correlated stocks are selected using our proposed SSDTW algorithm, which takes into account both DTW similarity and TLCC.  
\end{itemize}  

The results in Table~\ref{AblationStudy} demonstrate that the SSDTW algorithm achieves the best performance, followed by the DTW algorithm, while the RS method yields the poorest results in predicting future stock prices. This finding highlights that the performance improvement is not simply due to an increase in input dimensionality but rather the effective selection of correlated stocks. Furthermore, while stocks selected by the DTW algorithm provide some benefits, their contribution is limited as they fail to capture the TLCC essential for accurate stock price prediction.

\subsection{Sensitivity Analysis}
\label{sec: sensitive}

In this section, we conduct a sensitivity analysis on weather dataset to evaluate the impact of the hyperparameter $K^s$ in Eq.\ref{eq2}, $K^e$ in Eq.\ref{eq6} and $\lambda$ identified in Eq.\ref{eq7} and Eq.\ref{eq8} on the overall performance of the proposed model. This analysis aims to understand how variations in these hyperparameter influence the model's predictions and stability.

To further investigate the impact of the number of selected correlated sequences $K^s$ mentioned in Eq.\ref{eq2}, we perform a sensitivity analysis by varying the number of correlated sequences from 0 (direct prediction) to 1, 3, 5, and 10. The experimental results in Figure.~\ref{fig:Sen_CNT} consistently indicate that selecting 3-5 correlated sequences achieves optimal performance across both MSE and MAE metrics.

This finding can be attributed to two key factors. Selecting fewer than three correlated sequences may fail to provide sufficient contextual information, potentially overlooking critical interdependencies. On the other hand, incorporating too many sequences may introduce noise and irrelevant signals, which can adversely affect the model's predictive accuracy. 

In addition, we also analyze the impact of selecting different numbers of positive and negative samples $K^e$ during the training of CLE on the performance of using CLE to select correlated sequences. We trained the CLE with various sample ratios (1:1, 3:3, 5:5, 3:10, and 10:3) and applied these trained CLEs to select correlated sequences for predictions across multiple models. The results in Figure.~\ref{fig:Sen_posneg} indicate that, compared to direct predictions, training CLE with these varying sample ratios significantly enhances prediction accuracy for nearly all models, demonstrating a high degree of robustness in the framework's predictions, with consistent MSE and MAE results.

Finally, we analyze the sensitivity of the regularization parameter $\lambda$ identified in Eq.\ref{eq7} and Eq.\ref{eq8}. By varying $\lambda$ from 0.2 to 0.5, 0.7, and 0.9, we observe a minimal impact on the model's generalization capability across most models. The results presented in Figure.~\ref{fig:Sen_lambda} indicate that the performance remains stable across this range of $\lambda$ values, demonstrating a high degree of robustness in the framework's predictions, with consistent MSE and MAE results.

\subsection{Case Study}
To more intuitively demonstrate the effectiveness of the proposed SSDTW algorithm and the CLE optimization method in selecting sequences with TLCC relationships, we use the sequence with city id = 1066 as the target sequence on the real estate dataset with relatively short sequence lengths. Relevant sequences are selected respectively according to the SSDTW algorithm and the CLE method, and the results are shown in Figure.\ref{fig:case-study}. The price sequence with id 4276 exhibits a very clear TLCC correlation with the target sequence, and both sequence selection methods are able to correctly identify this sequence. We also observe that the sequence with the highest similarity computed by the TLCCSP framework does not necessarily exhibit an obvious TLCC correlation. This further corroborates our conclusion in Section.~\ref{sec: sensitive} that multiple sequences should be selected as references in prediction to achieve optimal performance.

\section{Conclution}

In this paper, we present TLCCSPP, a novel framework that leverages the SSDTW algorithm to identify and utilize time-lagged cross-correlations (TLCC) among sequencess in various domains. By capturing these complex inter-sequence relationships, the framework enhances the accuracy of deep-learning based predictions. To improve computational efficiency, we propose a contrastive learning-based encoder that approximates SSDTW distances, significantly reducing computational overhead. This makes the TLCCSP framework more practical for real-world applications. 

Experimental results on weather, finance and real estate time series datasets demonstrate the effectiveness of our framework.
On the weather dataset, SSDTW reduces MSE by 16.01\% compared with single-sequence methods, while the CLE further decreases MSE by 17.88\%. On the stock dataset, SSDTW achieves a 9.95\% MSE reduction, and CLE reduces it by 6.13\%. For the real estate dataset, SSDTW and CLE reduce MSE by 21.29\% and 8.62\%, respectively. Additionally, the contrastive learning approach decreases SSDTW computational time by approximately 99\%.

\section{GenAI Disclosure Statement}

In this study, we utilize OpenAI's GPT-4.1 model solely for the purpose of grammar checking and light editing of the manuscript. We confirm that we do not employ generative AI technologies in other critical aspects of the research, such as code development, data analysis, or experimental design.

We acknowledge that while generative AI serves as a valuable tool for editing and language enhancement, it does not replace the critical thinking and originality required in the research process.

\bibliographystyle{ACM-Reference-Format}
\bibliography{main}

\end{document}